\documentclass[letterpaper, 10 pt, conference]{ieeeconf/ieeeconf}
\IEEEoverridecommandlockouts
\usepackage{graphics}
\usepackage{epsfig}
\usepackage{times}
\usepackage{amsmath}
\usepackage{amssymb}

\usepackage[hidelinks]{hyperref}
\usepackage{booktabs}
\usepackage{hhline}
\usepackage{color}
\usepackage[table]{xcolor}
\usepackage{comment}
\usepackage{multirow}
\usepackage{subcaption}
\usepackage[labelfont=bf, font={footnotesize}]{caption}

\newcommand{\colorin}[1]{
    \cellcolor[HTML]{FDFD96}{#1}
}

\title{\LARGE \bf
Data Augmentation for NeRFs in the Low Data Limit
}

\author{Ayush Gaggar and Todd D. Murphey
\thanks{The authors are with the Department of Mechanical Engineering at Northwestern University, Evanston, IL 60208 USA. Contact: \href{mailto:agaggar@u.northwestern.edu}{\tt\small agaggar@ u.northwestern.edu}, \href{mailto:t-murphey@northwestern.edu}{\tt\small t-murphey@northwestern.edu}.}
\thanks{This material is supported by the National Science Foundation, under Grant CNS-2237576.}
}

\begin{document}

\maketitle

\begin{abstract}
Current methods based on Neural Radiance Fields fail in the low data limit, particularly when training on incomplete scene data. 
Prior works augment training data only in next-best-view applications, which lead to hallucinations and model collapse with sparse data. 
In contrast, we propose adding a set of views during training by rejection sampling from a posterior uncertainty distribution, 
generated by combining a volumetric uncertainty estimator with spatial coverage. 
We validate our results on partially observed scenes; on average, our method performs 39.9\% better with 87.5\% less variability across established scene reconstruction benchmarks, as compared to state of the art baselines.
We further demonstrate that augmenting the training set by sampling from \emph{any} distribution leads to better, more consistent scene reconstruction in sparse environments.
This work is foundational for robotic tasks where augmenting a dataset with informative data is critical in resource-constrained, \emph{a priori} unknown environments. Videos and source code are available at \textcolor{blue}{\emph{\href{https://murpheylab.github.io/low-data-nerf/}{https://murpheylab.github.io/low-data-nerf/}}}
\end{abstract}

\section{Introduction}

Determining how to collect additional data with little prior information is critical in robotics, since robots need to plan and act in unknown environments.
Recently, Neural Radiance Fields (NeRFs) \cite{OriginalMildenhallNerf} have seen an explosion in research, largely due to their incredible performance in creating high quality, complex 3D scene reconstructions.
Indeed, NeRFs in robotics are being explored for object manipulation \cite{kerr2022evonerf, dai2023graspnerf, liu2024rgbgrasp}, mapping and SLAM \cite{slam_nerf_sucar2021imap, slam_nerf_Yan2023iccv, zhan2022activermap}, control \cite{control_adamkiewicz2022}, or with natural language processing \cite{wang2022clip, rashid2023language, kerr2023lerf}. 

Enabling embodied learning for NeRFs, where a robot iteratively and incrementally collects additional data, could greatly advance resource-constrained applications where photo-realistic renderings are needed. 
Despite the potential benefits of using robots to collect samples in environments challenging for humans, such as medicine, underwater, or outer space, applying NeRFs to these environments remains an open challenge.

\begin{figure}[t!]
\centering
\includegraphics[width=0.98\linewidth]{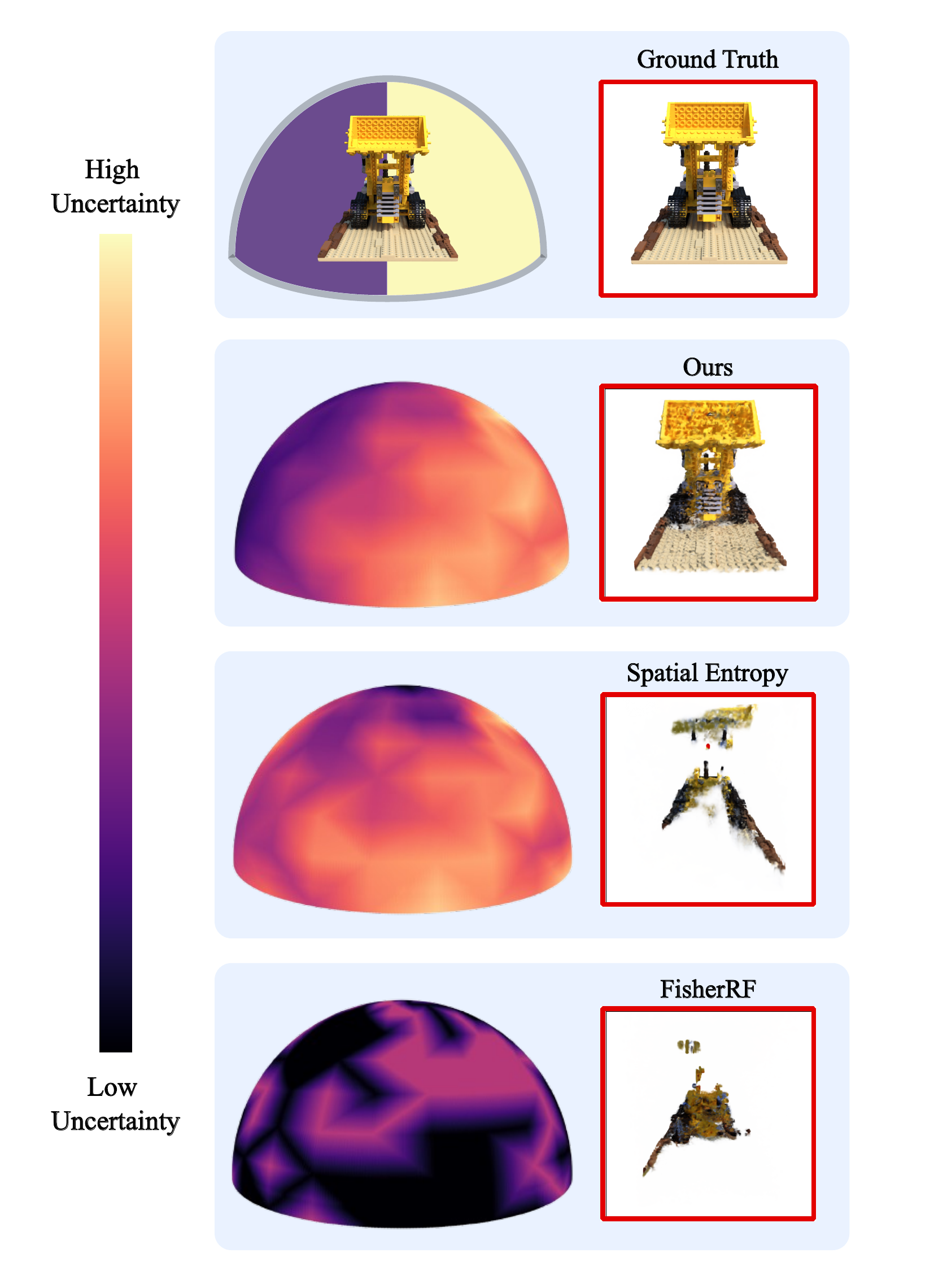}
\caption{\textbf{Uncertainty distributions (left) and scene reconstruction (right) for NeRFs with sparse training views.} 
Per method, \textit{Left}: Uncertainty distribution generated over the hemisphere bounding the object, with brighter colors corresponding to higher uncertainty. 
\textit{Right}: Novel view reconstruction after training with sparse images (6 initial + 6 augmented).
Comparisons with probabilistic methods---ours, Spatial Entropy \cite{lee2022uncertainty}, 
and FisherRF \cite{jiang2023fisherrf}---are shown.
\emph{Ground truth} shows that the initial views are all taken from the left half of the hemisphere, such that the object is only partially observed and the right half is highly uncertain.
After augmenting the training set, our method does the best job in both accounting for unseen regions in the uncertainty distribution field, and in scene reconstruction quality. 
On the other hand, Spatial Entropy has high uncertainty even in observed regions (indicative of hallucinations), and FisherRF has low uncertainty across most the hemisphere, even in unobserved regions (indicative of overfitting).}
\label{fig:hemispheres}
\vspace{-5pt}
\end{figure}

Like many other learning algorithms, the \textit{selection} of data for NeRFs matters significantly in the low data limit, as we see in Fig. \ref{fig:hemispheres}; with enough training data, curating the training set ceases to improve NeRF performance. 
Conversely, training NeRFs in resource-constrained scenarios with partial, sparse, or low-quality images can lead to models that collapse, hallucinate, or are overfit, as seen in Fig. \ref{fig:hallucinate}. 
When applying NeRF-based methods in realistic environments with humans, especially in robotics, an accurate understanding of the model's confidence in its own predictions, especially incorrect ones, is crucial. 

Data augmentation based on uncertainty currently focuses on information maximization \cite{lee2022uncertainty, jiang2023fisherrf, catnips}, modifying the NeRF architecture \cite{pan2024many, sunderhauf2023density, pan2022activenerf} or fully covering the scene \cite{est3dUncField, kopanas2023improving}. 
Research in NeRFs with sparse inputs largely focuses on architecture \cite{yu2020pixelnerf, yang2023freenerf}, pre-trained networks \cite{ViewFormer, DietNeRF}, or additional sensors \cite{kangle2021dsnerf, SparseNeRF}, rather than data augmentation.
None of these methods provide a framework to add a batch of views beyond generalizing next-best-view, nor do they explicitly deal with partially observed scenes.

Our work addresses the current lack of techniques for augmenting NeRFs trained on little available data.
Specifically, we select a \emph{set} of images to add to the training set by rejection sampling from a posterior distribution, which accounts for both in and out of distribution uncertainty. 
Each experiment is set up such that only half the object is initially observed with a sparse set of training views ($N=$ 6 images). 
We demonstrate that our method substantially outperforms state-of-the-art data selection methods in terms of scene reconstruction. 
Furthermore, we show that regardless of the particular choice of uncertainty estimation, sampling from the chosen posterior distribution results in better, more consistent performance. 
Our work trains end-to-end with no pre-trained networks nor offline learning and can easily be adapted to existing NeRF architectures with minimal change or overhead.

\section{Related Work}
In this section, we focus our literature review to work on active view selection for radiance field data augmentation, as well as NeRFs trained on sparse inputs. We encourage readers to refer to \cite{GaoKyle2023NNRF} for overviews on NeRF literature, \cite{WangGuangming2024NiRA} for various robotic applications of NeRFs, \cite{RenPengzhen2022ASoD} for active deep learning, and \cite{AL_uncertainty} for uncertainty quantification in deep learning.

\subsection{Active Learning in NeRFs}\label{subsec:nbv}
Next-best-view (NBV) selection is a fundamental area of research in robotics; however, active learning with NeRFs has been restricted to selecting views that \emph{most maximize} their measure of uncertainty, without considering a sequence or set of views.
In determining NBV, data augmentation methods generally fall under the three categories we explore below.

\subsubsection{Greedy, Information Maximization Methods}
Lee et al. \cite{lee2022uncertainty} maximize entropy of the weight distribution along candidate views; we call this method ``Spatial Entropy" and compare against it in our experiments. 
ActiveNeRF \cite{pan2022activenerf} and NeRF-W \cite{martin2021nerf} modify the NeRF architecture to output variance, Yan et al. \cite{yan2023active} model variability in the network weights, and BayesRays \cite{goli2023} and FisherRF \cite{jiang2023fisherrf} (a method we compare against) maximize Fisher Information. All of these methods select the top $N$ views with the maximum variance, entropy, or information gain.
Wilkinson et al. \cite{wilkinson2024adaptive} plans a trajectory, but instead of statistically rejecting, simply reject views if the metric performance is below a threshold, similar to the metric proxy used by Ran et al. \cite{ran2023neurar}.

By using the learned model itself, information maximization methods do a great job of quantifying uncertainty for in-distribution (ID) views the model has been trained on. For out-of-distribution (OOD) views though, these methods are either unreliable or incorrectly confident.

\begin{figure}[t!]
\vspace{6pt}
        \centering
        \captionsetup{labelfont=small}
        \begin{subfigure}[b]{0.475\linewidth}
            \includegraphics[width=\linewidth]{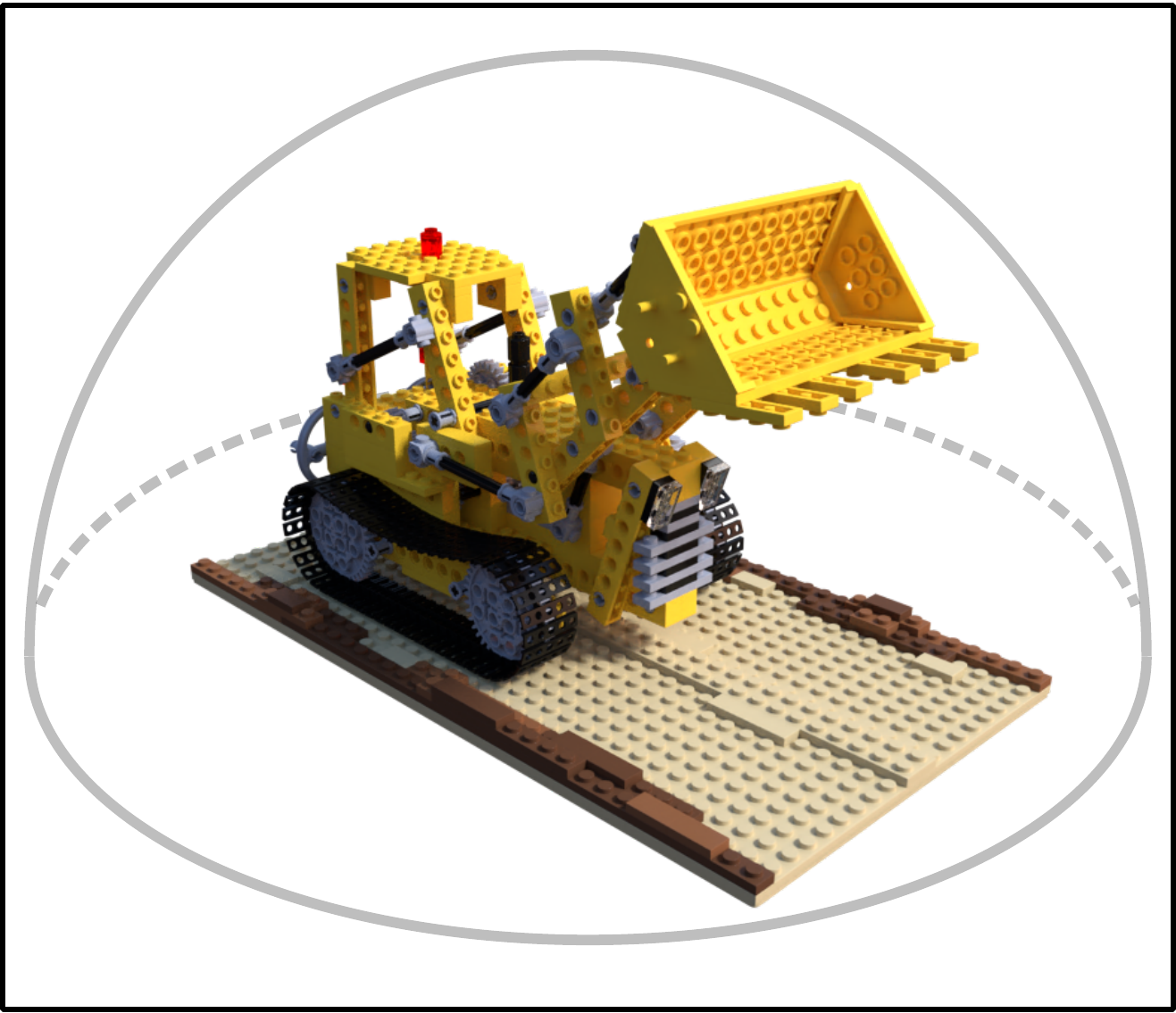}
            \caption{\small Lego Bulldozer Object}
        \end{subfigure}
        \hfill
        \begin{subfigure}[b]{0.475\linewidth}
            \includegraphics[width=\linewidth]{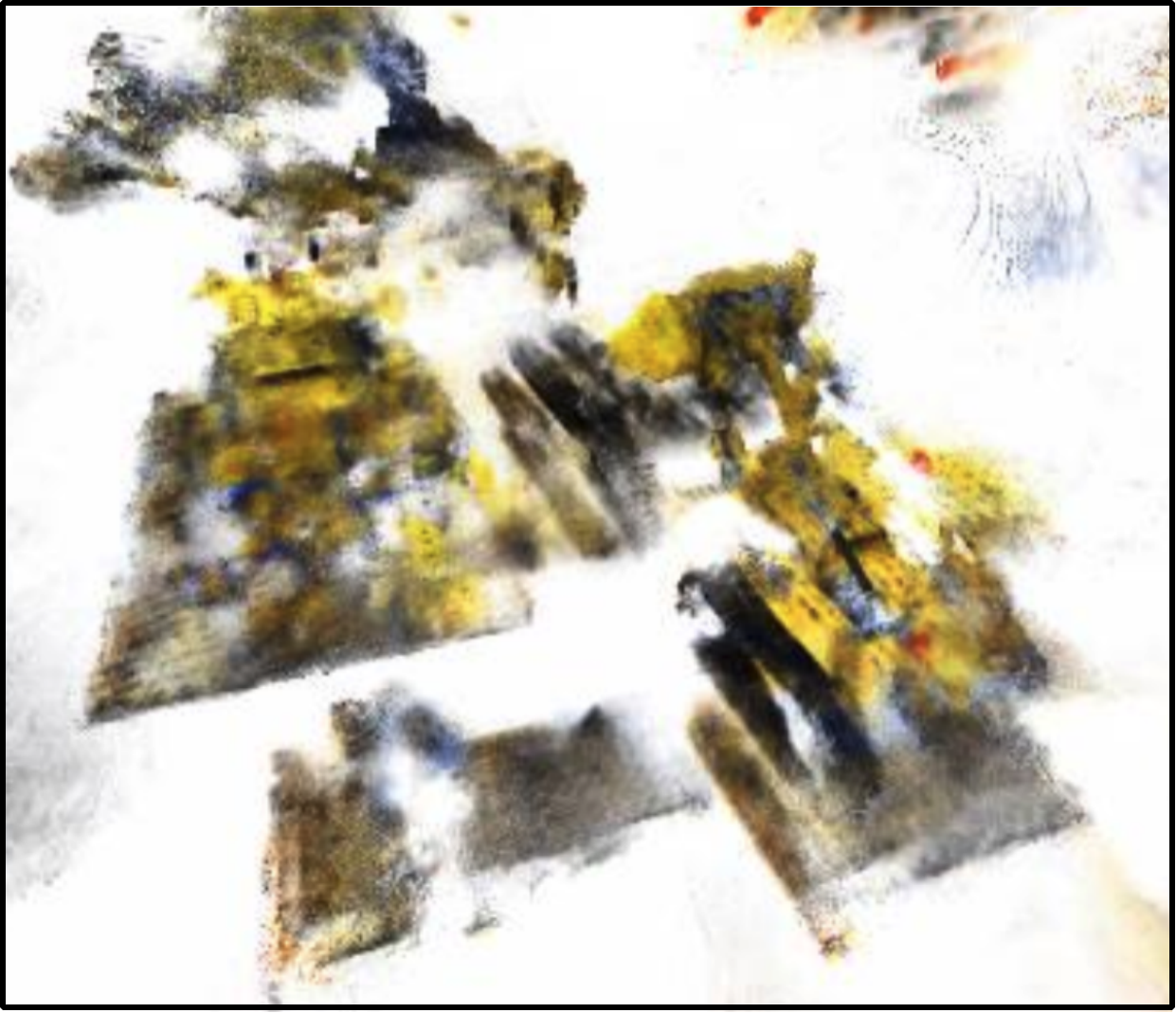}
            \caption{\small Failure by Hallucination}
            \label{fig:hallucinate-overfit}
        \end{subfigure}
        \vskip0.2\baselineskip
        \begin{subfigure}[b]{0.475\linewidth}
            \includegraphics[width=\linewidth]{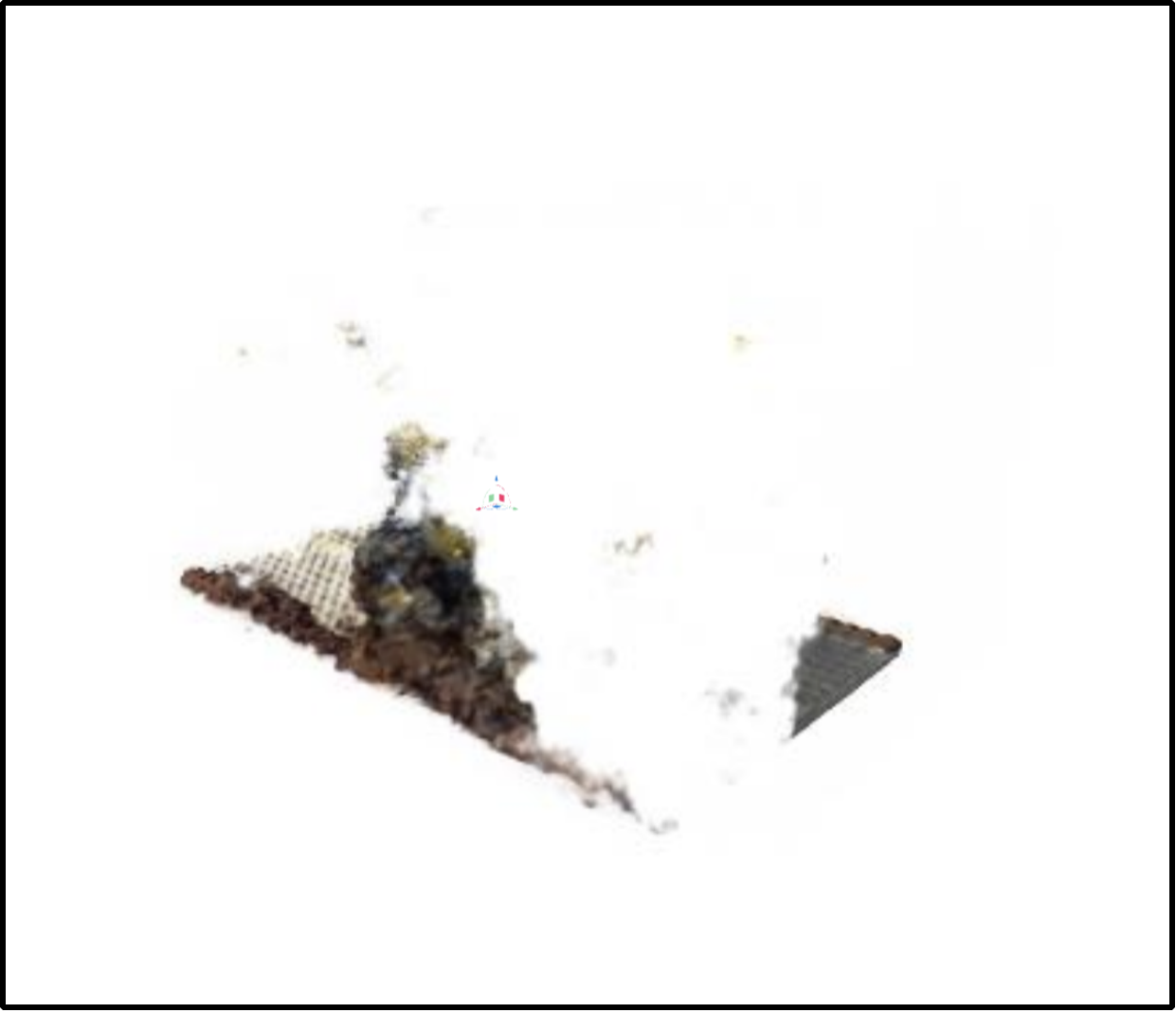}
            \caption{\small Failure by Overfitting}
        \end{subfigure}
        \hfill
        \begin{subfigure}[b]{0.475\linewidth}
            \includegraphics[width=\linewidth]{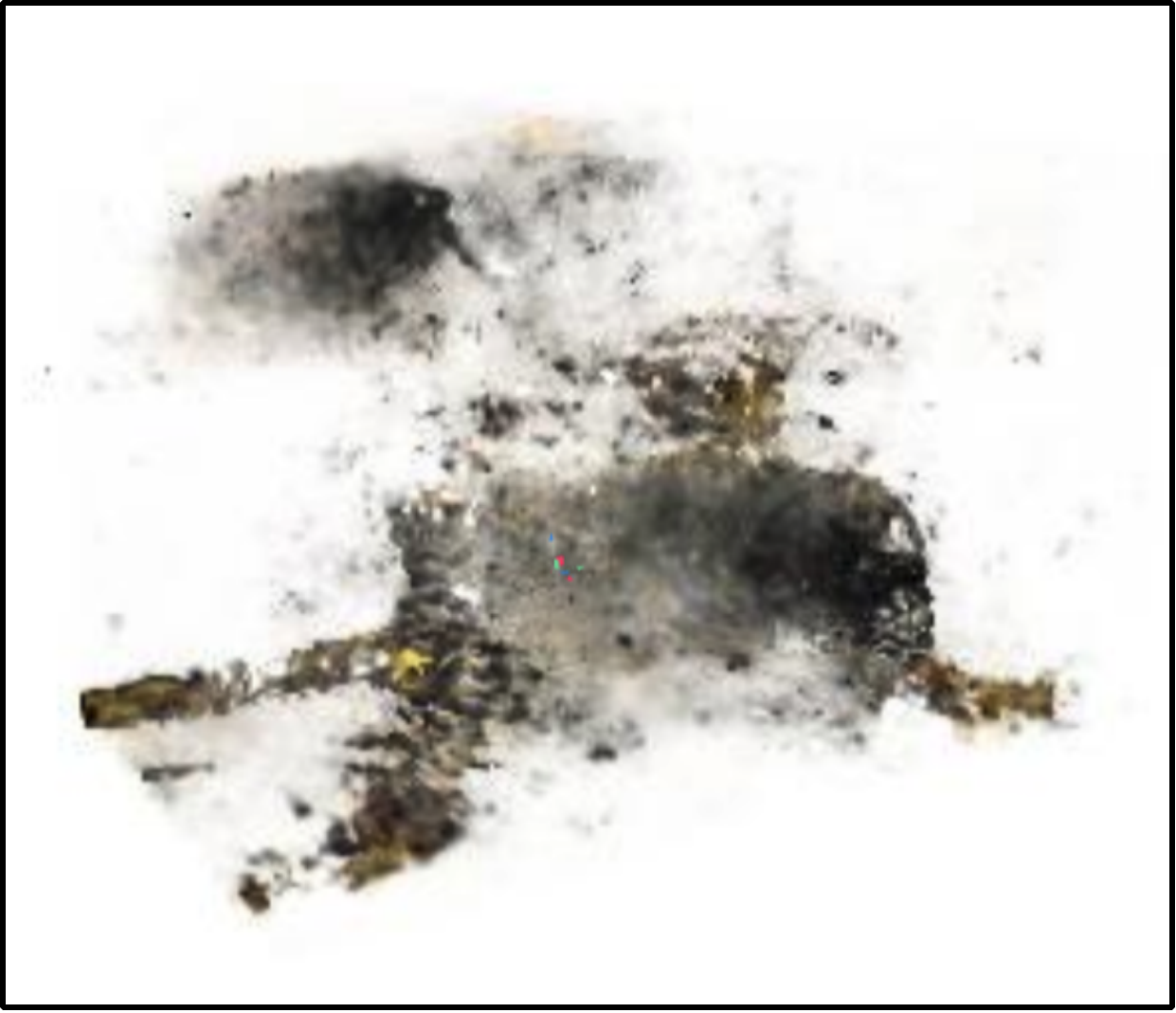}
            \caption{\small Failure by Blurriness}
            \label{fig:hallucinate-blurry}
        \end{subfigure}
    \vspace{2pt}
    \captionsetup{labelfont=bf}
    \caption{\textbf{Common ways NeRFs fail in the low data limit.} 
    (a) Training views can be added along the hemisphere, outlined in gray; here, the Lego bulldozer object is shown. (b) Failure by hallucination, where rays are unable to learn depth properly and fail to create a single model; data augmented by the FisherRF method. (c) Failure by overfitting, where the model confidently predicts nothing in the scene; data augmented by the Spatial Entropy method. (d) Failure by occluded artifacts, where the model is unable to render with clarity; data augmented by adding views furthest from each other, i.e., maximally apart.}
\label{fig:hallucinate}
\vspace{-5pt}
\end{figure}

\subsubsection{Geometric Methods}
Mildenhall et al. \cite{OriginalMildenhallNerf}, MipNeRF \cite{barron2022mip}, and work derived from \cite{OriginalMildenhallNerf} add views sampled from a uniform distribution between [0, $\frac{\pi}{2}$] and [0, 2$\pi$], forming a hemisphere; we call this method ``Uniform", and compare it against our method.
Shen et al. \cite{est3dUncField} and Xue et al. \cite{xue2024neural} evaluate occluded and unseen regions to place additional camera views, while other scene-based techniques \cite{kopanas2023improving} incorporate heuristics to maximize distance between training views. If training images are too spatially spread out though, DS-NeRF \cite{kangle2021dsnerf} shows that NeRF performance suffers from visual artifacts.

Scene-based methods prioritize scene coverage in neural rendering, which overcomes the downfalls of the previously discussed greedy methods. Although they do a far better job in quantifying OOD uncertainty, they fail to differentiate uncertainty between ID detailed areas and occluded or unexplored areas. 

In contrast, our method combines the benefits of both information-driven and scene coverage methods, and thus accurately classifies both challenging details and unexplored regions as high uncertainty. 
 
\subsubsection{Augmenting with Deep Learning}
Pan et al. \cite{pan2024many} create labels using ShapeNet \cite{chang2015shapenet}, NeU-NBV \cite{jin2023neu} uses a long short-term memory (LSTM) module, and Sünderhauf et al. \cite{sunderhauf2023density} ensemble NeRFs to model uncertainty. 
S-NeRF \cite{S-NeRF} and CF-NeRF \cite{CF-NeRF} use Variational Inference by sampling from a latent space, estimating a distribution over the parameters of all possible radiance fields.
Ensemble learning is an attractive, Bayesian approach to modeling uncertainty, but comes at the expense of computation or complex architectural changes.
On the other hand, not only does our method train from scratch, but it's also easily integrated into any NeRF architecture.

\subsection{NeRFs with Sparse Inputs}
\label{subsec:sparse_in}
PixelNeRF \cite{yu2020pixelnerf} was the first to train NeRFs with sparse inputs, but scaled poorly with each additional training image. 
Sin-NeRF \cite{xu2022sinnerf}, ViewFormer \cite{ViewFormer}, and DietNeRF \cite{DietNeRF} use pre-trained Transformers to guide training, and NeRFDiff \cite{gu2023nerfdiff} uses a pre-trained diffusion model. SparseNeRF \cite{SparseNeRF} adds depth data in addition to RGB to overcome visual blurriness with sparse inputs. FreeNeRF \cite{yang2023freenerf} and RegNeRF \cite{niemeyer2022regnerf} both improve the NeRF architecture itself for sparse inputs.

In our experiments with sparse images, we discover that training views cannot be spatially too diverse or too similar, otherwise the model collapses or has visual artifacts, as shown in Fig. \ref{fig:hallucinate}. 
It is well known in NeRF literature that training views should be both plentiful and diverse \cite{pan2024many, kangle2021dsnerf, nerf-drawbacks}, but little research is done on what views suffice, let alone how to algorithmically select them.

Here we lay a foundation for using robotics in NeRFs with sparse inputs, as robots can simply collect additional data based on areas with high uncertainty. 
Rather than just extrapolating NBV techniques, we use rejection sampling as outlined in \cite{doucet1998sequential} to augment the training dataset with a \emph{sequence} of images. Moreover, our work explicitly reasons about partially observed scenes and trains from scratch.

\section{Methodology}
\subsection{Preliminary - NeRF Architecture}
The original NeRF architecture \cite{OriginalMildenhallNerf} models a 3D scene as a continuous function using a multi-layer perceptron (MLP) to output color $\mathbf{c}$ = $(r, g, b)$ and volume density $\mathbf{\sigma}$ as a function of Cartesian position $\mathbf{x}$ = $(x, y, z)$ and 2D viewing angles $\mathbf{v}$ = $(\theta, \phi)$. 
NeRF uses volume rendering to generate the color of the pixel from a target viewing angle and position, called a camera ray. Given a camera ray
\begin{math}
    r(s) = o + sd, 
\end{math}
where \begin{math}
    o \in \mathcal{R}^3
\end{math}
represents the center of the camera, $d$ is the direction vector, and $s$ represents how far along the camera ray to view, the expected color of a camera ray is formulated as:
\begin{equation}\label{eq:nerf_color}
        C(r(s)) = \int_{s_n}^{s_f}T(s)\sigma(r(s))\textbf{c}(r(s), \mathbf{v})ds,
\end{equation}
where \begin{math}
    T(s) = \textrm{exp}(-\int_{s_n}^{s_f}\sigma(r(s'))ds')
\end{math}. $T(s)$ denotes the accumulated transmittance along the ray from the near bound ${s_n}$ to the far bound ${s_f}$ of the scene. In practice, this integral is approximated via the quadrature rule with $R$ spaced bins along the ray $r(s)$.

In this work, we build off the NeRF architecture of the NerfStudio team \cite{nerfstudio}, as many recent advances are incorporated into a single python package. 

\begin{figure*}[tp!]
    \centering
    \includegraphics[width=\textwidth]{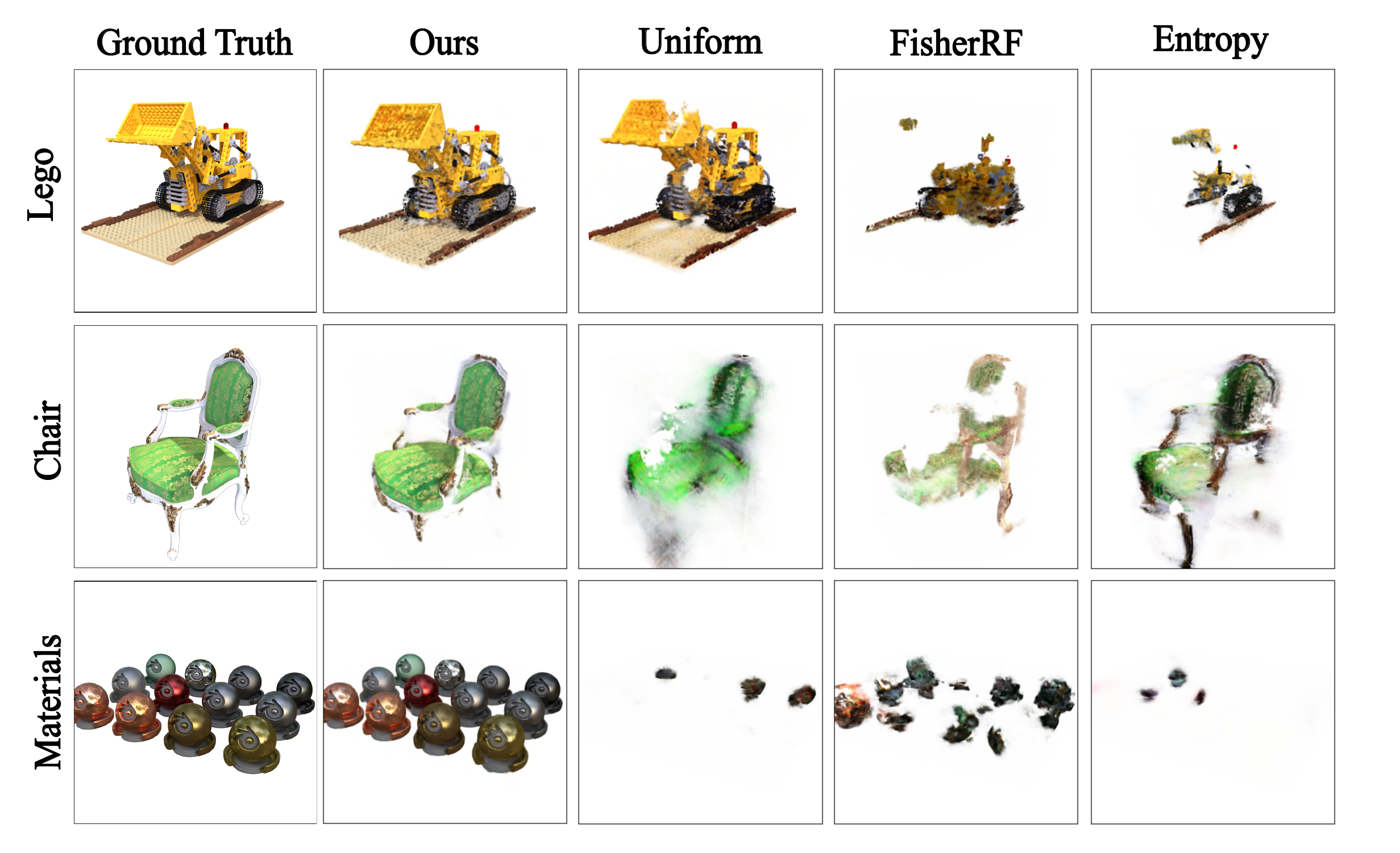}
    \caption{\textbf{Scene reconstruction after 10k training iterations for three different objects and data augmentation methods.} Across all scenes, only our method renders the model without visual artifacts. 
    The scene is initially partially observed, with six training views all taken from the same half of the hemisphere; based on the data selection method, six additional views are added to the training set after 200 training iterations.}
    \label{fig:qualitative-comp}
    \vspace{-5pt}
\end{figure*}

\subsection{Uncertainty Quantification}
Uncertainty quantification in NeRFs needs to account for both ID and OOD rays to account for both insufficient and partial data respectively.
As motivated in Subsec. \ref{subsec:nbv}, model-based methods excel at quantifying ID uncertainty while scene-based methods can capture views in unseen areas. With sparse images, each term individually is too myopic to properly capture uncertainty; thus, we model predictive uncertainty by summing up the two uncertainty terms, the first of which quantifies ID uncertainty through entropy and the second of which encourages spatial coverage:
\begin{equation}\label{eq:total-uncertainty}
    \mathbf{U(r(s))} = \mathbf{H(r(s))}_{ent} + \mathbf{D(r(s))}_{dist}
\end{equation}
Our uncertainty equation can be calculated spatially over the scene, since it is a function of camera rays $r(s)$.

\subsubsection{In-Distribution Uncertainty}
The first term takes the entropy described in \cite{lee2022uncertainty}, which can be interpreted as the uncertainty in the probability of a ray $r(s)$ to traverse the length of the scene without encountering an object. Specifically,
$\mathbf{H(r(s))}$ represents the Shannon Entropy at $r(s)$:
\begin{equation}\label{eq:spatial-entropy}
    \mathbf{H(r(s))}_{ent} = -\sum_{i=1}^{R} w_i(r(s)) \log_2(w(r(s)))
\end{equation}
Here, $w(s)$ represents the derivative of the opacity function, which determines the probability with which a given ray intersects the surface of an object:
\begin{math}
    w(s) = \frac{dO(s)}{ds} = \frac{d(1 - T(s))}{ds} = T(s)\sigma(s).
\end{math}
This is often calculated by NeRF architectures as a discrete approximation of \begin{math}
    w_i = T_i(1-\exp({-\sigma_i(s_{i+1} - s_i)})
\end{math}, where the summation to $R$ represents stratified sampling along the ray $R$ times by amount \begin{math}
    s \in [0, 1]
\end{math}. 
Since $w(s)$ is a valid probability density function, computing its Shannon Entropy allows for per-pixel ID uncertainty quantification. 

\subsubsection{Out-of-Distribution Uncertainty}
Greater scene coverage exposes the model to more views of the object, decreasing the number of OOD rays.
Therefore, we propose hemisphere coverage as a suitable proxy to quantify uncertainty for OOD rays, as unseen regions will be further from current training views.
The second term represents the total $L^2$-distance between existing views and a candidate ray:

\begin{equation}\label{eq:dist-to-viewpoints}
    \mathbf{D(r(s))}_{dist} = \sum_{n=1}^{N}f(g_n, g_{r(s)}),
\end{equation}
where $f$ is the $L^2$-distance function for Lie groups, $g$ is an SO(3) rotation matrix, $g_n$ is the SO(3) representation of one of $N$ training views in the training dataset, and $g_{r(s)}$ is the SO(3) representation at the candidate ray $r(s)$.
We convert the viewing angles $\mathbf{v}$ to rotation matrices $g$ in SO(3) and apply the equivalent L2-distance in Lie groups, as defined by Fan et al. \cite{fan2017online}:
\begin{equation}\label{eq:lie-dist-SO3}
    f(g_1, g_2)=\frac{1}{2} \| \log(g_1^{-1} g_2) \|_M^2,
\end{equation}
where \begin{math}
    \|x \|_M = \sqrt{x^TMx}
\end{math} 
and \begin{math}
    M \in \mathbb{R}^{n \times n}
\end{math} is symmetric positive definite, i.e., the identity matrix. Since we are fixing candidate rays to the hemisphere, which can be represented in SO(3), $\mathbf{D(r(s))}$ can be calculated using Eq. \ref{eq:lie-dist-SO3}.

By summing up the entropy and spatial coverage terms, we create an uncertainty distribution over the hemisphere that overcomes the disadvantages of a purely model-based or geometry-based understanding of the scene, while capturing both ID and OOD uncertainty.

\subsection{Sampling from a Distribution}
Augmenting the training set by selecting only the most informative views can lead to overfitting the model, as seen in Fig. \ref{fig:hallucinate-overfit}. Similarly, prioritizing only coverage can lead to blurriness, as seen in Fig. \ref{fig:hallucinate-blurry}. 
Unlike selecting the maximal value, sampling from a distribution can improve performance by overcoming local minima and increasing the diversity of total samples.
Although well explored in robotics with particle filters, Monte Carlo methods, and importance sampling, little research is done for collecting samples for NeRFs. 
Rejection sampling specifically accepts samples based on a probability proportional to the target density, allowing for a richer set of samples that better represent the uncertainty distribution.
Therefore, instead of greedily selecting the top $N$ views, we choose to sample $N$ views from our distribution.

The total uncertainty calculated by Eq. \ref{eq:total-uncertainty} can be evaluated per candidate ray $r(s)$ with viewing angles $\mathbf{v}$.
Since NeRF papers conventionally bound scenes on a hemisphere of known radius and height, candidate viewing angles $\mathbf{v}$ can be extrapolated to a pose in SE(3) with known rotation matrix and position in $(x, y, z)$ space.

We reject or accept $N_{new}$ viewpoints based on standard rejection sampling \cite{doucet1998sequential}, which can generate samples with density proportional to the uncertainty distribution. A summary of the rejection sampling procedure is as follows:
\begin{enumerate}
    \item Sample a new ray $r_{\bar{x}}$ from the hemisphere, as per \cite{OriginalMildenhallNerf}.
    \item Calculate $w = \textbf{U($r_{\bar{x}}$)}$ from Eq. \ref{eq:total-uncertainty}.
    \item Sample \begin{math}
        u \sim \mathcal{U}_{[0, 1]}
    \end{math}.
    \item Accept $r_{\bar{x}}$ if \begin{math}
        u \leq w / M
    \end{math}, where $M$ is a scalar, typically $\sim [1, 20]$, to scale the proposal distribution.
    \item Repeat 1-4 until $N_{new}$ views are added.
\end{enumerate}

\begin{figure*}[htp!]
\centering
\includegraphics[width=\textwidth]{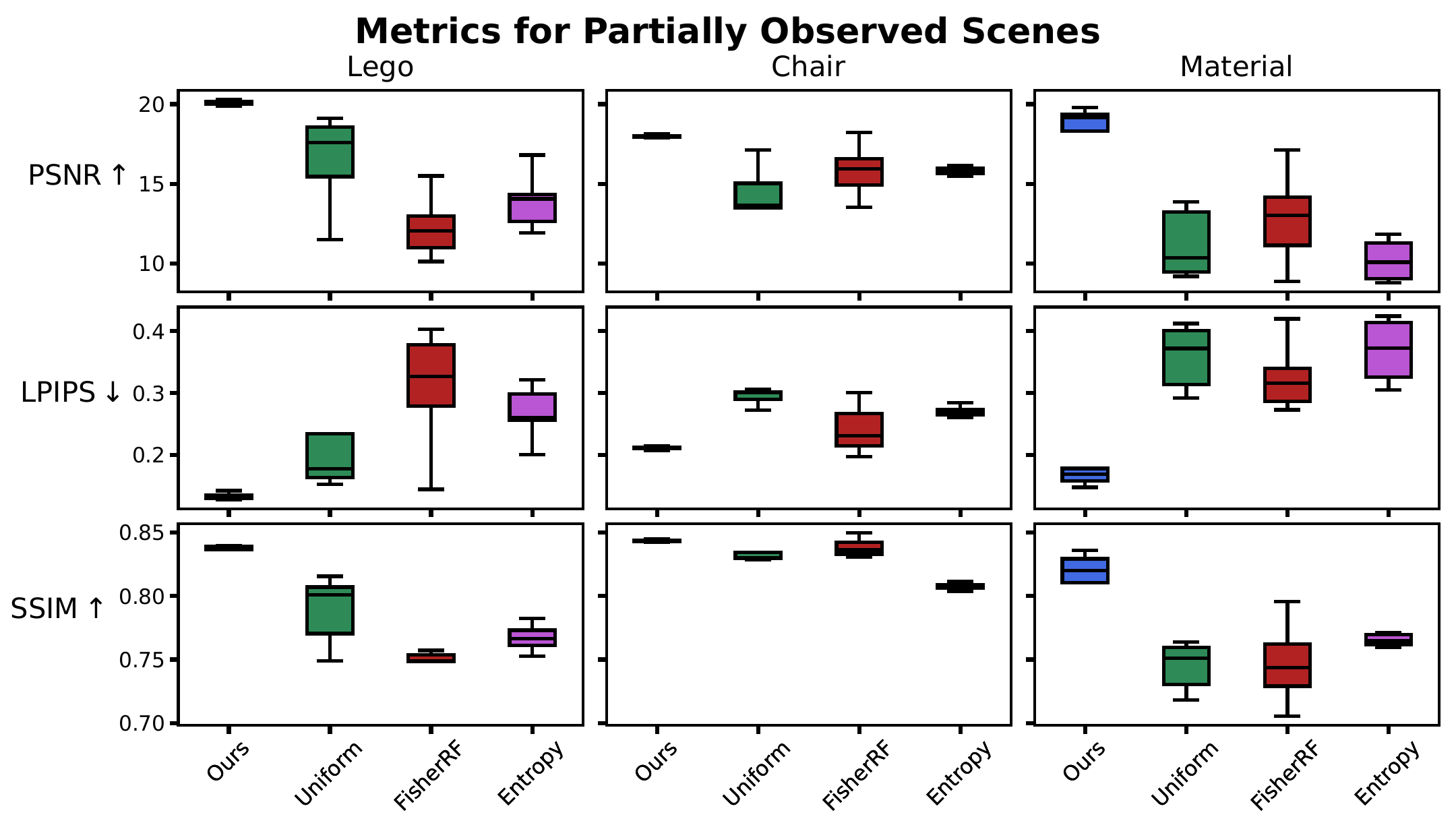}
\caption{\textbf{Evaluation results of standard image quality metrics across our method and three other SOTA baselines.} Each metric score was evaluated across the 200 images in the evaluation dataset for each of the three scenes. A higher score is better for PSNR and SSIM, and a lower score is better for LPIPS. 
We achieve the best median performance and the lowest interquartile range compared to any method across each scene, except for material SSIM vs. \emph{Entropy}. Our method performs better with a statistical significance of p \textless 0.05 and a Bonferroni correction of 3, 
except for lego LPIPS vs. \emph{Uniform}, chair LPIPS vs \emph{FisherRF}, and chair SSIM vs \emph{Uniform and FisherRF}.}
\label{fig:metrics-comp}
\vspace{-5pt}
\end{figure*}

\section{Experiments}
\subsection{Problem Statement}
Embodied agents in unknown environments with no prior data can easily collect a sparse set of samples locally, after which exploration is needed to gather information about the scene. We set up a similar experiment in simulation, where the initial 6 views are taken from one half of the hemisphere, such that half the object is never seen before. 6 additional images are collected based on the different data selection methods. 
Our main goal is to augment a sparse dataset to result in the best novel view reconstruction, specifically when initial views do not comprehensively cover the object. 

\subsection{Setup}
\textbf{Datasets.}
We evaluate our approach using three scenes from the standard Blender dataset \cite{OriginalMildenhallNerf}---Lego bulldozer, chair, and materials. The treads and wheels of the bulldozer, the thin legs of the chair, and the different textures and details on the material orbs make each scene uniquely difficult to render. We follow the evaluation protocol of \cite{OriginalMildenhallNerf}, where metrics are averaged over 200 evaluation images with a resolution of 800 x 800 and fixed camera intrinsics.

\textbf{Scene.} 
All the scenes are rendered in Blender with camera poses facing inward on a hemisphere, following \cite{OriginalMildenhallNerf}. The object can be bounded by a box of known dimensions, such that it is at the center of the hemisphere. The initial six images are selected such that only half the scene is initially visible and are the same images per trial for each object. For all benchmarks, six additional images are taken after 200 iterations, and training is done for 10,000 iterations.

\textbf{Metrics.} Performance is measured using three standard image quality metrics---peak signal-to-noise ratio (PSNR), structural similarity index (SSIM), and LPIPS \cite{zhang2018perceptual}. A higher PSNR score indicates better image quality, a lower LPIPS score denote images that are closer to the ground truth, and a higher SSIM score indicates better perceptual quality. The final metric per trial is the mean of the metric computed for all 200 evaluation images.

\textbf{Baselines.} We compare our method of data selection with three leading baselines, Uniform \cite{OriginalMildenhallNerf}, FisherRF \cite{jiang2023fisherrf}, and Spatial Entropy (or just entropy) \cite{lee2022uncertainty}, of which the latter two are probabilistic. Along with being probabilistic, these baselines were chosen because they also require minimal changes to NeRF architecture. We incorporate the methods of data selection for each of these in Nerfstudio's Nerfacto architecture \cite{nerfstudio}, implemented in PyTorch, based on the code release or mathematical framework for each method.

\textbf{Implementation.} All hyperparameters used are set as the default Nerfacto options, with two prominent exceptions. First, the appearance embedding layer is removed in order to allow for variable dataset sizes (i.e., in order to go from six to twelve images after the first 200 iterations). Second, the learning rate in the ``Nerfacto Field'' is decreased from $0.01$ to $0.002$ for the chair and materials dataset for model stability when adding data. 

It takes around 11 minutes for our method and around 8 minutes for the others to train 10k iterations per scene on a single Nvidia RTX 6000 GPU; however, we note that our augmentation process could be easily parallelized.

All comparisons are run with identical architectures, hyperparameters, and initial scene setup, and data is added at 200 iteration steps for all scenes and methods. The initial weights of the model are randomly initialized for each of the 10 trials per scene. The only difference between the comparisons is the \emph{method} used to add six additional views to the training dataset.

\subsection{Results}
Figure \ref{fig:qualitative-comp} provides examples of performance comparison from each data collection method, and Fig. \ref{fig:metrics-comp} shows  quantitative results of different methods across the ten trials. 
Only our method in Fig. \ref{fig:qualitative-comp} renders a comprehensible model per scene, with no visual artifacts, blurriness, or occlusions. 
Although \textit{Uniform} does surprisingly well on the Lego bulldozer, it renders only a couple of the material orbs, potentially because detail for the orbs are visible only in the narrow band of viewing angles low to the ground. Both \textit{FisherRF} and \textit{Entropy} struggle similarly to render the entire object for either of these scenes, likely because both are model-based augmentation methods.
In Fig. \ref{fig:metrics-comp}, we can see that our method has the best median performance across all scenes and methods.
\textit{Uniform} shows high variability in performance, likely because it can select samples from regions already covered by the initial images. \textit{FisherRF} and \textit{Entropy} also show significant variability, which may stem from overfitting causing some trials to fail to render OOD views.

Our method consistently outperforms the others across all three metrics---PSNR, LPIPS, and SSIM. Across the lego, chair, and materials scenes respectively, our method achieves a 66.71\%, 31.81\%, and 90.06\% higher median PSNR, 59.78\%, 29.85\%, and 54.70\% lower median LPIPS, and 11.88\%, 4.45\%, and 10.29\% higher SSIM scores than other methods. Furthermore, our method yields far tighter interquartile ranges;
across the lego, chair, and materials scenes respectively, our method has 94.95\%, 94.47\%, and 71.99\% less PSNR variability, 94.43\%, 95.42\%, and 77.31\% less LPIPS variability, and 93.29\%, 89.00\%, and 76.37\% less SSIM variability.

These results underscore the effectiveness of sampling from Eq. \ref{eq:total-uncertainty}, leading to consistently superior performance. They further demonstrate that our uncertainty term better captures both in-distribution and out-of-distribution uncertainty.

\subsection{Ablation Study}
Metrics are evaluated on the lego bulldozer scene, with median and interquartile range (IQR) shown for $N=10$ trials. Higher median score corresponds to better scene reconstruction, and lower IQR corresponds to decreased variability across trials.

\textbf{Effect of individual uncertainty terms.}\label{subsubsec:ind-terms}
Our proposed uncertainty, Eq. \ref{eq:total-uncertainty}, sums up spatial entropy for ID uncertainty and $L^2$-distance for OOD uncertainty.
We study the influence of these two terms by ablating them; Table \ref{tab:terms-ablation} compares augmenting based on solely maximizing entropy or maximizing distance from the training views.
Augmenting based on just entropy (ID uncertainty) does surprisingly well, far better than adding the furthest views (OOD uncertainty), likely because ID uncertainty is more important when half the scene is already seen; this trend is unlikely to continue if even less of the scene was initially visible. Still, the sum of both terms combined consistently yields the best metric, underscoring the importance of accounting for both ID and OOD uncertainty.

\textbf{Effect of rejection sampling.}\label{subsubsec:rej-sample}
We investigate the claim that adding a set of views from \textit{any} uncertainty distribution is better than generalizing NBV for NeRFs trained with sparse data. 
Table \ref{tab:sampling-ablation} shows that regardless of the data selection method, \emph{Sampling} from the uncertainty distribution (instead of just selecting the most uncertain view) generally improves results and reliability for that method. Particularly for \textit{Entropy}, sampling significantly boosted performance and reliability. 
As expected, our method still outperforms sampling from either of the other selection methods. These results show the effectiveness of sampling in overcoming myopia of data, regardless of the uncertainty method sampled from for data augmenting.

\section{Conclusions, Limitations, \& Future Work}
We have shown that the key to augmenting NeRFs for sparse, partially observed scenes lies in sampling from a distribution that accounts for both model-based, in-distribution uncertainty, along with scene-based, out-of-distribution uncertainty. 
Our proposed method significantly and reliably outperforms state of the art baselines in scene reconstruction. 
Furthermore, we demonstrate that sampling can improve NeRF performance, regardless of the method used to create the uncertainty distribution. 
An advantage of our work is that it can be seamlessly incorporated into most NeRF algorithms, requiring no architecture changes, offline training, or pre-trained networks.

\begin{table}[tp!]
\renewcommand{\arraystretch}{1.3}
\centering
\resizebox{\columnwidth}{!} 
    {\begin{tabular}{|c| c|c| c|c| c|c|}
        \hline
        \multirow{2}{*}{Method} & \multicolumn{2}{c|}{PSNR $\uparrow$ } & \multicolumn{2}{|c|}{LPIPS $\downarrow$} & \multicolumn{2}{|c|}{SSIM $\uparrow$} \\
        \cline{2-7}
        & \multicolumn{1}{c|}{Median} & IQR & \multicolumn{1}{|c|}{Median} & IQR & \multicolumn{1}{|c|}{Median} & IQR \\
        \hline
        Entropy Only & 17.9307 & 1.9793 & \multicolumn{1}{|c|}{0.1607} & 0.0323 & \multicolumn{1}{|c|}{0.8143} & 0.0185 \\
        \hline
        Distance Only & 12.3819 & 2.0699 & \multicolumn{1}{|c|}{0.3110} & 0.0627 & \multicolumn{1}{|c|}{0.7536} & 0.0206 \\
        \hline
        Combined & \multicolumn{1}{c|}{\colorin{20.0666}} & \colorin{0.1575} & \multicolumn{1}{|c|}{\colorin{0.1313}} & \colorin{0.0055} & \multicolumn{1}{|c|}{\colorin{0.8379}} & \colorin{0.0025} \\
        \hline
    \end{tabular}}
\vspace{2pt}
\caption{\textbf{Ablation study --- Impact of ID (entropy) vs OOD (distance) terms from Eq. \ref{eq:total-uncertainty}.} Data is added based on just entropy, just distance, or combined. Our method achieves both significantly better performance and reliability across all three metrics of scene reconstruction. Results are for $N=10$ trials.
The best performing method is highlighted in yellow.}
\label{tab:terms-ablation}
\end{table}

\begin{table}[tp!]
\renewcommand{\arraystretch}{1.3}
\setlength{\extrarowheight}{2pt}
\centering
\resizebox{\columnwidth}{!} 
    {\begin{tabular}{|c|c| c|c| c|c| c|c|}
        \hline
        \multicolumn{2}{|c|}{\multirow{2}{*}{Method}} & \multicolumn{2}{|c|}{PSNR $\uparrow$ } & \multicolumn{2}{|c|}{LPIPS $\downarrow$} & \multicolumn{2}{|c|}{SSIM $\uparrow$} \\
        \cline{3-8}
        \multicolumn{2}{|c|}{} & \multicolumn{1}{|c|}{Median} & IQR & \multicolumn{1}{|c|}{Median} & IQR & \multicolumn{1}{|c|}{Median} & IQR \\
        \toprule[0.5pt]
        \hline
        
        \multirow{2}{*}{Ours} & NBV & \multicolumn{1}{|c|}{16.9309} & 3.8392 & \multicolumn{1}{|c|}{0.2025} & 0.0957 & \multicolumn{1}{|c|}{0.7884} & 0.0362 \\
        \hhline{~-------}
        & Sampling & \multicolumn{1}{|c|}{\colorin{20.0666}} & \colorin{0.1575} & \multicolumn{1}{|c|}{\colorin{0.1313}} & \colorin{0.0055} & \multicolumn{1}{|c|}{\colorin{0.8379}} & \colorin{0.0025} \\
        \toprule[0.5pt]
        \hline
        
        \multirow{2}{*}{FisherRF} & NBV & \multicolumn{1}{|c|}{12.0366} & 2.1618 & \multicolumn{1}{|c|}{0.3265} & 0.0993 & \multicolumn{1}{|c|}{0.7490} & \colorin{0.0051} \\
        \hhline{~-------}
        & Sampling & \multicolumn{1}{|c|}{\colorin{13.0377}} & \colorin{1.9886} & \multicolumn{1}{|c|}{\colorin{0.2988}} & \colorin{0.0722} & \multicolumn{1}{|c|}{\colorin{0.7685}} & 0.0242 \\
        \toprule[0.5pt]
        \hline
        
        \multirow{2}{*}{Entropy} & NBV & \multicolumn{1}{|c|}{14.0598} & \colorin{1.7082} & \multicolumn{1}{|c|}{0.2599} & 0.0427 & \multicolumn{1}{|c|}{0.7663} & 0.0185 \\
        \hhline{~-------}
        & Sampling & \multicolumn{1}{|c|}{\colorin{17.9307}} & 1.9793 & \multicolumn{1}{|c|}{\colorin{0.1607}} & \colorin{0.0323} & \multicolumn{1}{|c|}{\colorin{0.8143}} & \colorin{0.0121} \\
        \hline
    
    \end{tabular}}
\vspace{2pt}
\caption{\textbf{Ablation study --- Sampling from a distribution vs. taking NBV evaluation metrics.} Here we compare the performance of each probabilistic method with itself, when adding data based on NBV selection vs. when sampling from a distribution. Sampling from each respective distribution achieves significantly better median metric performance for scene reconstruction regardless of the method, and generally decreases variability in model performance.
Results are for $N=10$ trials. The best performing method is highlighted in yellow.}
\label{tab:sampling-ablation}
\vspace{-5pt}
\end{table}

Like other works, our work assumes that the position and size of the object are bounded by a hemisphere of known size.
In the future, a more general approach could be developed with robotic hardware where the object's position and size are both unknown \textit{a priori}. We are also interested in dynamic trajectory optimization, considering scene coverage, robot dynamics, and energy to move to new viewpoints when controlling a robotic arm for scene exploration.

\section*{Acknowledgment}
The authors would like to acknowledge Allison Pinosky, Muchen Sun, and Joel Meyer for their insights during this project. Any opinions, findings, conclusions, or recommendations expressed in this material are those of the authors.

\pagebreak
\addtolength{\textheight}{-16cm}
\bibliographystyle{IEEEtran}
\bibliography{IEEEabrv,references.bib}

\end{document}